# Digital divides in scene recognition: Uncovering socioeconomic biases in deep learning systems

Michelle R. Greene[1], Mariam Josyula[2], Wentao Si[3], Jennifer A. Hart[4]

1: Barnard College, Columbia University (mgreene@barnard.edu)
2: University of Pennsylvania (mjosyula@seas.upenn.edu)
3: Dartmouth College (wentao.si.gr@dartmouth.edu)
4: Bates College (jenniferhart11@gmail.com)

Corresponding author: Michelle R. Greene (mgreene@barnard.edu)

# Abstract

Automatic scene classification has applications ranging from urban planning to autonomous driving, yet little is known about how well these systems work across social differences. We investigate explicit and implicit biases in deep learning architectures, including deep convolutional neural networks (dCNNs) and multimodal large language models (MLLMs). We examined nearly one million images from user-submitted photographs and Airbnb listings from over 200 countries as well as all 3320 US counties. To isolate scene-specific biases, we ensured no people were in any of the photos. We found significant explicit socioeconomic biases across all models, including lower classification accuracy, higher classification uncertainty, and increased tendencies to assign labels that could be offensive when applied to homes (e.g., “slum”) in images from homes with lower socioeconomic status. We also found significant implicit biases, with pictures from lower socioeconomic conditions more aligned with word embeddings from negative concepts. All trends were consistent across countries and within the diverse economic and racial landscapes of the United States. This research thus demonstrates a novel bias in computer vision, emphasizing the need for more inclusive and representative training datasets.

# Introduction

In recent years, the rapid advancement of computer vision technology has transformed many domains, ranging from autonomous vehicles to medical diagnostics. However, computer vision systems are not immune to human society's biases and inequalities. While there is increasing understanding that there are race, gender, and age-based gaps in computer detection of people from photographs and videos [1,2], as well as biased assumptions about groups of people from photographs [3–8], comparatively little is known about the biases that may exist for places. As scene-based information is now being used in various urban planning [9] and home valuation [10,11] contexts, assessing potential social biases in the computer vision classification of places is necessary.

Potential for bias exists in all steps of a standard computer vision pipeline. Deep learning architectures, including deep convolutional neural networks (dCNNs) and multimodal large language models (MLLMs) are known for their extensive data requirements. Engineers rely on web scraping to amass a broad sample of images for training purposes. Creating a scene database might begin with compiling an exhaustive list of scene categories and then employing web scraping to gather image exemplars for each category [12]. However, the label set can often contain offensive terms or terms that can be used in an offensive way [13–15], leading to the inclusion of problematic content. Sourcing images from the web has been shown to return images that exacerbate gender and racial biases [3]. Crowdsourcing is then used to ensure that the scraping procedure leads to correct category labels. Consequently, crowd workers' prejudices tend to be mirrored and even intensified in the final datasets [16].

Our study examines the biases inherent and latent in algorithmic scene classification. We consider three datasets totaling nearly one million images from real-world homes, comprising user-submitted home photographs and pictures from the Airbnb website. To ensure that any biases were scene-based, we verified that all photos were free of people. We examined both pretrained dCNNs and a state-of-the-art multimodal large language model (MLLM), GPT-4V (OpenAI) for explicit biases in classification accuracy, confidence, and specific classifications that could be offensive in the context of one's home (e.g., "slum"). Additionally, we examined implicit biases in Contrastive Language-Image Pre-Training (CLIP) [17], whereby images from lower socioeconomic status (SES) locations were more strongly aligned with language embeddings from negative concepts. Our results revealed stark explicit and implicit biases across models and datasets. Pretrained dCNNs were less accurate and less confident in classifying images of homes from lower SES while simultaneously showing a heightened tendency to assign potentially offensive labels. Unfortunately, these biases did not disappear with a MLLM. This pattern persisted not only in global comparisons but also within the varied economic and racial landscapes of the United States.

These findings illuminate the profound influence of socioeconomic and developmental factors on computer vision performance, highlighting significant concerns about unintentional biases in automated image categorization systems. Such biases underscore the pressing need for more inclusive and representative training datasets to mitigate these disparities and ensure fairer outcomes in AI applications.

# Results

## Explicit biases in user-submitted images

We first considered a set of ~1200 home images submitted by families in 54 countries on the Dollar Street website [18]. This site, maintained by the nonprofit organization Gapminder, was created to provide visual representations of how families live across different socioeconomic circumstances worldwide. We evaluated classification accuracy (top-1 and top-5) using six deep convolutional neural networks (dCNNs) that were either pretrained or fine-tuned on the Places database [12] to classify scenes. All findings revealed a notable disparity in classification accuracy compared to the established benchmark from the Places test set; see Table 1 for the pretrained networks and Supplementary Table 1 for all networks. These results demonstrate that user-submitted images are not classified as accurately by these networks as images obtained from standard Internet datasets.

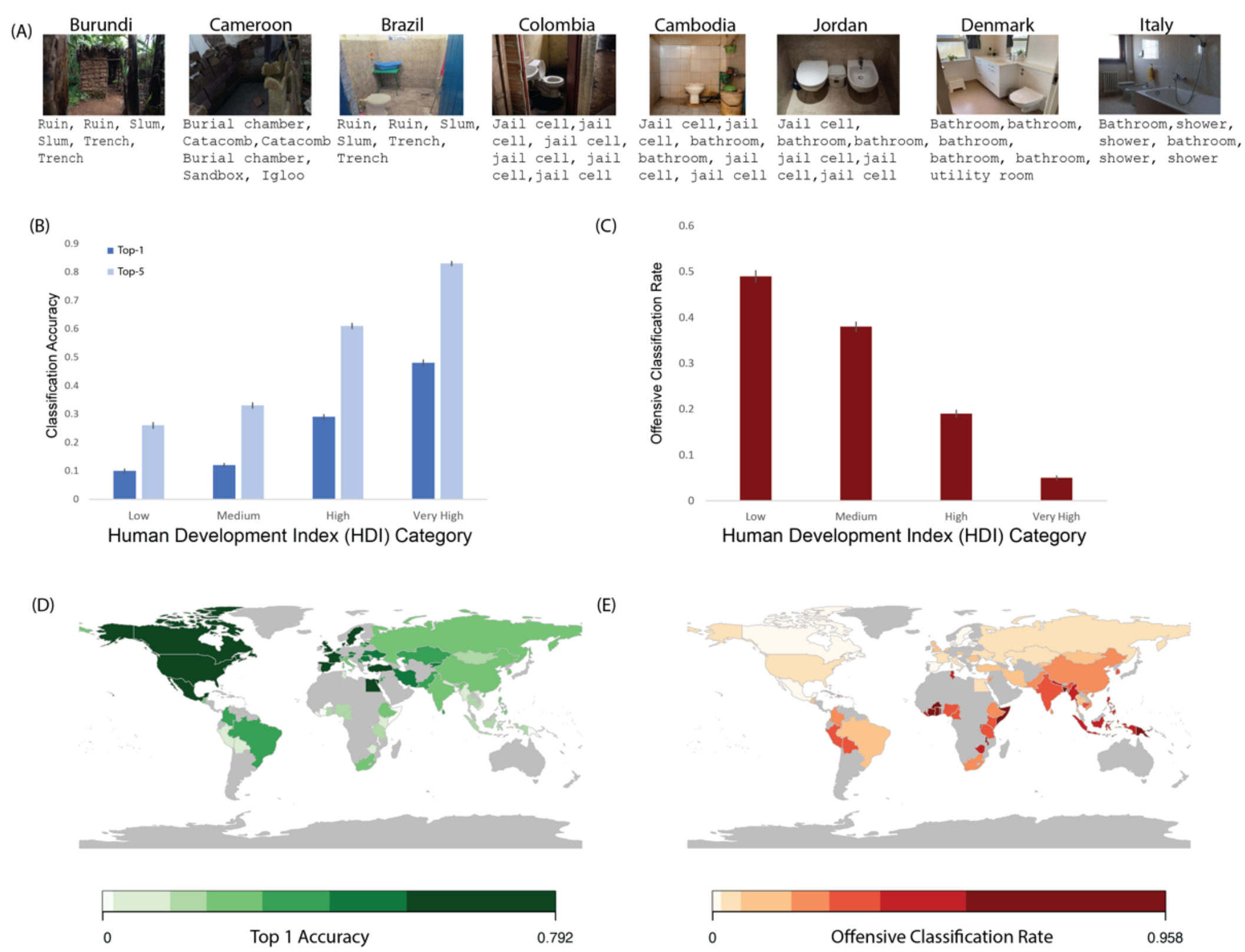

**Figure 1: (A) Examples of bathroom images from Dollar Street with sample classifications from the six dCNNs (order: Alexnet, Resnet-18, Resnet-50, Resnet-101, Resnet-152, ConvNext. Other categories included bedrooms, kitchens, and living rooms. (B) Top-1 and top-5 accuracy as a function of Human Development Index (HDI) category. (C) Offensive classification rate as a function of HDI category. (D) Map of top-1 accuracy, averaged across the six dCNNs. (E) Map of offensive label rate, averaged across the six dCNNs. All error bars represent 95% confidence intervals.**

| Network | Metric | Places benchmark | Dollar Street | $X^2$ |
|---|---|---|---|---|
| Alexnet | Top-1 | 53% | 21% | 485.7 |
| Resnet-18 | Top-1 | 55% | 32% | 268.9 |
| Resnet-50 | Top-1 | 55% | 33% | 229.2 |
| Alexnet | Top-5 | 83% | 47% | 1094.3 |
| Resnet-18 | Top-5 | 85% | 59% | 627.9 |
| Resnet-50 | Top-5 | 85% | 62% | 508.3 |

**Table 1: Performance on Dollar Street image set versus Places test set benchmarks. For all networks and both top-1 and top-5 accuracy, all differences are highly statistically significant ($p<<0.001$).**

**Pretrained dCNNs are less accurate in classifying low-SES places.** To assess the dCNNs for potential SES bias, we employed a linear mixed-effects modeling approach, predicting classification accuracy (top-1 and top-5) from family income and Human Development Index (HDI) category as fixed effects, with individual images and network architectures treated as random effects. Given the wide monthly income range ($27 to $14,753), family income was log-transformed for analysis. The total explanatory power of the models was notable, with the top-1 model achieving a conditional $R^2$ of 0.64 with a marginal $R^2$ of 0.43 and the top-5 model a conditional $R^2$ of 0.74 with a marginal $R^2$ of 0.52.

We observed that higher family income significantly predicted increased accuracy in both top-1 (beta: 1.16, $p<0.0001$) and top-5 (beta: 1.55, $p<0.0001$) classifications. This suggests that dCNNs are impaired when classifying images from less affluent homes. Additionally, a country's HDI category strongly predicted classification accuracy, with a monotonic increase in the log odds of accurate classification as HDI categories increased (Cochran-Armitage trend test Top-1: $Z=-27.7$, $p<0.0001$; Top-5: $Z=-36.8$, $p<0.0001$), see Figures 1B and 1D). These results indicate that dCNNs are less capable of classifying images from countries with lower economic and social development.

**Pretrained dCNNs are less confident in classifying low-SES places.** We extended these findings by considering classification uncertainty indexed by classification entropy. We found that family income and HDI had moderate explanatory power for classification entropy (conditional $R^2$ of 0.39 with a marginal $R^2$ of 0.09). Higher family income was associated with lower classification entropy (-0.26, 95% CI [-0.32, -0.21]), suggesting that dCNNs had more classification confidence for homes from wealthier households. A linear regression confirmed the monotonic relationship between HDI category and classification entropy ($F(1, 7140)=296.4$, $p<<0.0001$; $\beta$=-0.27 95% CI [-0.24 -0.30]). Together, these results mirror the picture we observed for classification accuracy, showing higher confidence for more affluent families and countries.

Critically, we found a robust negative association between classification entropy and accuracy. Specifically, logistic regression models were employed to predict both top-1 and top-5 accuracies based on classification entropy. The results indicated that entropy was a significant negative predictor for both top-1 accuracy (beta = -0.73, 95% CI [-0.78, -0.68], $p<0.001$) and top-5 accuracy (beta = -0.68, 95% CI [-0.72, -0.64], $p<0.001$). Furthermore, an aggregation of mean classification accuracies by continent demonstrated extremely high negative correlations with classification entropy (top1: $r=-0.99$, $t(2)=-10.2$, $p<0.01$; top5: $r=-0.99$, $t(2)=-8.25$, $p<0.05$). This pattern suggests that classification entropy can be a robust proxy for classification accuracy, particularly in contexts where ground truth category labels are unavailable. Additionally, these findings underscore the utility of classification entropy as it is a continuous measure, enabling detailed item-wise analyses that can provide granular insights into the comparative successes and failures of dCNNs.

**Pretrained dCNNs are more likely to classify low-SES places in offensive ways.** Nine scene categories from the Places dataset were designated as potentially offensive for their associations with disrepair, decay, incarceration, or death (burial chamber, catacomb, cemetery, jail cell, junkyard, landfill, mausoleum, ruin, slum). Although offensive terms are most commonly associated with people rather than places, scholars have noted that terms such as “slum” are not universally defined [19–21] and thus serve to marginalize people in poverty [19,22]. As this manuscript concerns images of private homes, each of these nine terms is unambiguously inappropriate. We noted whether these terms were included in the top five network classifications. Overall, 27% of images contained offensive classifications. We employed a generalized linear mixed-effects modeling approach to predict the probability of an offensive label using family income and HDI as predictors, with network and image as random effects.

The model's total explanatory power is substantial (conditional $R^2 = 0.62$, with a marginal $R^2$ of 0.38 due to the fixed effects alone). We observed a strong inverse relationship between family income and the likelihood of offensive classifications; higher income substantially reduced these potentially problematic classifications (-0.90, $p<0.001$). Similarly, a country's HDI category significantly influences the odds of an offensive label. Relative to countries with a 'very high' HDI, the log odds of receiving an offensive label increased monotonically with decreasing HDI levels ('high': 0.86, $p=0.003$; 'medium': 1.46, $p<0.001$; 'low': 1.50, $p<0.001$, see Figure 1C and 1E). This monotonic trend was confirmed with a Cochran-Armitage trend test (Z=30.5, p<0.0001).

**A multimodal large language model (MLLM) is less accurate in classifying low SES places.** We employed GPT-4V (OpenAI) to classify each image to assess potential bias in more state-of-the-art models. As with the dCNNs, we assessed classifications for correctness and potential offensiveness. We fit a linear mixed model to predict correctness from family income and Human Development Index (HDI) category as fixed effects, with individual images as random effects. The model's explanatory power was moderately high (marginal $R^2$=0.46, with 0.44 attributable to the fixed effects). As with the dCNNs, we observed a strong relationship between family income and classification accuracy (beta = 1.04, 95% CI [0.80, 1.27], p < .001), indicating that photos from wealthier families were more likely to be correctly classified by GPT-4V. While each country's HDI category did not emerge as a statistically significant factor when family income was taken into account, we observed a monotonically increasing classification accuracy across HDI categories (low: 43%, medium: 56%, high: 79%, very high: 91%) and this linear trend was confirmed with a Cochran Armitage test (Z=13.3, p<<0.0001).

**A multimodal large language model (MLLM) makes more potentially offensive classifications in low SES locations.** As with the dCNNs, we observed a strong inverse relationship between family income and the likelihood of offensive classifications; higher income substantially reduced these potentially problematic classifications (beta=-5.2, p<<0.0001). While each country's HDI category was not significant after considering family income, a Cochran-Armitage trend test demonstrated a strong linear trend in classification accuracy as a function of HDI (low: 3%, medium: 2%, high: 0.6%, very high: 0%, Z=3.1, p=0.002). We see that, like the older dCNN architectures, modern MLLMs are more accurate in classifying more affluent homes and are less likely to classify them in offensive ways.

Many of GPT-4V's classifications were vague (e.g., "room", "dwelling"). Further, the vague classification rate declined with increasing family income (beta=-0.83, p<<0.0001) and with each country's HDI category (low: 37%, medium: 34%, high: 15%, very high: 5%, Z=-10.2, p<0.0001). This pattern of equivocation could suggest that this MLLM employs more superordinate terms when it has less classification confidence or could result from an overt debiasing policy [23] whereby potentially offensive classifications are avoided by providing a vague, neutral term.

**Predicting dCNN misclassifications with country attributes.** What socioeconomic and demographic attributes might explain these patterns of relative success in deep learning models? We conducted a series of analyses using linear mixed models, predicting accuracy (top-1 and top-5), classification confidence, and offensive label rates as a function of factors from the CIA World Factbook: area, GDP per capita, population, literacy rate, percent urbanization, percent of GDP from tourism, infant mortality rate, education expenditure, youth unemployment rate, life expectancy at birth, maternal mortality rate, median age, and net migration rate. For each model, countries and network architectures were treated as random effects. All results are listed in Table 2. The explanatory power of all models was high, with conditional $R^2$ ranging from 0.81-0.93 and marginal $R^2$ between 0.15-0.60. We found that images from wealthier and more urban countries were classified more accurately, confidently, and with fewer offensive terms than those from poorer, more rural countries. Conversely, images from countries with lower infant and maternal mortality and higher life expectancy at birth were better classified compared to less healthy countries.

| Measure | CR2, MR2 | Intercept | GDP per capita | Population | Literacy | Urbanization | Tourism | Infant mortality | Education expense | Unemployment | Life expectancy | Maternal mortality | Median Age | Area | Net migration |
|---|---|---|---|---|---|---|---|---|---|---|---|---|---|---|---|
| Top-1 | **0.81**<br>**0.42** | -0.01 | **0.09***** | 0.004 | 0.03 | **0.07***** | 0.02 | -0.04 | 0.03 | -0.01 | **0.07***** | **-0.05**** | **0.05*** | 0.02 | -0.008 |
| Top-5 | **0.93**<br>**0.52** | 0.09 | **0.11***** | 0.02 | **0.08**** | **0.11***** | 0.04 | **-0.07*** | 0.04 | -0.02 | **0.09***** | **-0.09***** | **0.08**** | 0.01 | -0.004 |
| Entropy | **0.85**<br>**0.15** | **2.86**** | **-0.14**** | -0.006 | -0.006 | **-0.13*** | -0.02 | 0.06 | -0.04 | -0.04 | -0.08 | **0.16**** | **-0.10*** | 0.02 | -0.004 |
| Offense | **0.89**<br>**0.60** | 0.55 | **-0.07***** | -0.003 | **-0.08***** | **-0.08***** | **-0.04*** | **0.07***** | -0.02 | 0.02 | **-0.06***** | **0.06***** | **-0.09***** | -0.02 | 0.003 |

**Table 2: Results of linear mixed model analyses for DollarStreet dataset with dCNNs. Beta values are listed in the cells from column 3 onward. Note: predictors were whitened prior to model fitting. Thus, the sign and relative magnitudes are meaningful, but the coefficient values do not straightforwardly translate to specific units. (***: $p<0.001$; **: $p<0.01$; *: $p<0.05$).**

**Predicting MLLM misclassifications with country attributes.** We used the same predictors to assess the performance of GPT-4V. The results of these analyses confirmed the overall patterns found for the dCNNs. All regression coefficients are found in Table 3. For classification accuracy, our model had a conditional $R^2$ of 0.36, with 0.27 attributable to the fixed effects. This model predicted higher accuracy for countries with higher GDP per capita, literacy, urbanization, and life expectancy, with lower predicted accuracy for countries with higher infant and maternal mortality rates. Our model of offensive classification rate had an adjusted $R^2$ of 0.36. This model indicated that offensive classifications increased with increasing infant and maternal mortality rates and decreased with increasing literacy and life expectancy. Finally, we modeled the proportion of vague classifications as a function of socioeconomic and demographic characteristics. This model had a conditional $R^2$ of 0.33, with 0.24 associated with the fixed effects. We found that GPT-4V increased its rate of vague classification with increasing infant and maternal mortality and decreased its vague classification with increasing GDP per capita, urbanization, and life expectancy. These findings suggest that GPT-4V exhibits systematic biases in classification that align with the socioeconomic and demographic characteristics of the regions represented in the dataset and mirror those observed in deep convolutional neural networks (dCNNs).

| Measure | Intercept | GDP per capita | Population | Literacy | Urbanization | Tourism | Infant mortality | Education expense | Unemployment | Life expectancy | Maternal mortality | Median Age | Area | Net migration |
|---|---|---|---|---|---|---|---|---|---|---|---|---|---|---|
| Top-1 | -1.5 | **0.43**** | 0.002 | **0.35**** | **0.56***** | 0.16 | **-0.35**** | 0.08 | -0.15 | **0.64***** | **-0.34**** | **0.33*** | 0.09 | -0.16 |
| Offense | 0.008 | -0.0006 | -0.0006 | **-0.01**** | 0.005 | -0.001 | **0.014***** | -0.006 | 0.003 | **-0.013**** | **0.011**** | -0.0005 | -0.001 | 0.0007 |
| Vague | -0.57 | **-0.54**** | 0.08 | -0.23 | **-0.53***** | -0.05 | **0.32**** | -0.13 | 0.07 | **-0.51***** | **0.28*** | -0.28 | -0.19 | 0.11 |

**Table 3: Results of linear mixed model analysis for DollarStreet dataset with GPT-4V. Beta values are listed in the cells from column 2 onward. Note: predictors were whitened prior to model fitting. Thus, the sign and relative magnitudes are meaningful, but the coefficient values do not straightforwardly translate to specific units. (***: p<0.001; **: p<0.01; *: p<0.05).**

## Implicit biases in user-submitted images

Next, we tested a multimodal deep learning system to determine the extent of its implicit biases. Although GPT-4V still made potentially offensive classifications highly associated with low SES places, debiasing strategies might provide a pathway to mitigating explicit biases. However, these policies may cover rather than correct intrinsic biases.

To explore this, inspired by the implicit association test (IAT) from social psychology [24], we examined the associations between visual content and semantic descriptors using CLIP. Specifically, we analyzed whether image embeddings systematically aligned with words representing positive or negative concepts and whether these alignments varied as a function of socioeconomic indicators.

**Lower SES photographs are more strongly aligned with negative concepts.** We computed the mean alignment between each image and 20 positive and 20 negative concepts (see Methods for specific words). We then created a difference score by subtracting the negative alignment scores from the positive alignment scores. Thus, positive values show better alignment with positive concepts, while negative values indicate better alignment with negative concepts. To assess the extent to which this implicit bias score could be predicted from socioeconomic status, we fit a linear mixed effects model with the family's log income and the country's HDI category as fixed effects and country as a random effect. We found that family income was a significant predictor of positive bias (beta = 2.27e-03, 95% CI [1.87e-03, 2.67e-03], t(876) = 11.23, p < .001). While HDI was not significant when considering family income, we observed a monotonically increasing positivity with increased HDI (low: -0.0026, medium: -0.0014, high: 0.0005, very high: 0.0027) that was confirmed with a Jonckheere-Terpstra test (JT=145254, p<0.0001). This model had a conditional $R^2$ of 0.19, with 0.16 associated with the fixed effects. These results indicate that socioeconomic factors, particularly family income, strongly influence the implicit bias score. The overall proportion of explained variance is low, as might be expected given that many of the positive and negative concepts were not particularly visual (e.g., "peace", "laughter", "grief", and "stink"). However, the relatively high proportion of variance explained by fixed effects suggests that these socioeconomic variables are central to understanding the biases encoded in the model.

Next, we assessed which socioeconomic and demographic variables predict implicit bias in the CLIP model. We used a linear mixed model to predict the difference score from the CIA World Factbook variables. The model had moderate explanatory power (conditional $R^2$ of 0.20, with 0.09 associated with the fixed effects). Within this model, a country's per-capita GDP (beta = 1.26e-03, 95% CI [4.15e-04, 2.10e-03], t(1157) = 2.93, p =0.003) and life expectancy (beta =

1.53e-03, 95% CI [6.87e-04, 2.38e-03], $t(1157) = 3.56$, $p < .001$) were the only statistical effects. This indicates the wealthier and healthier countries were more strongly aligned with word vectors representing abstract positive concepts.

## Explicit biases in international Airbnb images

Our second dataset extended the previous findings to over 100,000 pictures sampled from 219 countries and regions on the Airbnb website. Unlike the DollarStreet set, which aims to show family life "as it is," the images on Airbnb are intended to advertise one's home to potential paying visitors. As such, this set provides a more rigorous test of potential socioeconomic scene biases. Ground-truth category labels were unavailable for this dataset, so we annotated a random sample of countries across the HDI spectrum to estimate classification accuracy. Given the small size of this sample, we will focus most of our analysis on classification entropy and the rate of potentially offensive labels.

**Pretrained dCNNs are less accurate in classifying low-SES places.** Using a set of ~1000 images sampled from 12 countries (three from each HDI category), we annotated ground truth categories and computed top-1 and top-5 classification accuracies. We fit linear mixed models predicting each accuracy type from HDI, using network architecture as a random effect. Our model of top-1 accuracy had a conditional $R^2 = 0.06$ with a marginal $R^2$ of 0.04 due to the fixed effects alone. HDI emerged as a strongly significant predictor of top-1 accuracy. Compared to the reference category of 'very high' HDI countries, we observed a monotonic increase in offensive labels with decreasing country development: 'high': beta = -0.30, 95% CI [-0.44, -0.16], p<0.001; 'medium': beta = -0.58, 95% CI [-0.73, -0.43], p < .001; 'low': beta = -1.30, 95% CI [-1.54, -1.06], p < .001. Similarly, top-5 classification accuracy decreased monotonically with decreasing HDI. Compared to 'very high' HDI countries, 'high': beta=-0.23, 95% CI [-0.42, -0.05], p<0.05; 'medium': beta=-0.49, 95% CI [-0.57, -0.20], p<0.001; 'low': beta=-0.81, 95% CI [-1.26, -0.81], p<0.001. Therefore, pretrained dCNNs are less capable of classifying homes from lower HDI locations, even when these homes are on public display, such as on Airbnb.

**Pretrained dCNNs are less confident in classifying low-SES places.** We applied a linear mixed model to evaluate the relationship of HDI category on classification entropy, treating network architecture as a random effect. The model had substantial explanatory power ($R^2$ of 0.71 and a marginal $R^2$ of 0.01 attributable to the fixed effects). As with the DollarStreet set, we observed a monotonic increase in classification entropy with decreasing HDI category. Compared to the baseline category of "very high" with a mean classification entropy of 3.09 bits, countries classified as "high" M=3.12, "medium" M=3.15, and "low" M=3.23. The monotonic entropy increase with decreasing HDI was confirmed with a Jonckheere-Terpstra test (JT=246770, p=0.003), indicating that lower HDI categories are associated with greater uncertainty in network classification. This trend aligns with findings from the DollarStreet set, reinforcing the relationship between socioeconomic factors and classification entropy.

**Pretrained dCNNs are more likely to classify low-SES places in offensive ways.** We determined that a country's HDI could predict the propensity of dCNNs to assign offensive labels to home images. The model demonstrated a significant explanatory power with a conditional $R^2$ of 0.25, with a marginal $R^2$ of 0.15 attributable to the fixed effects alone. Critically, low HDI emerged as a significant predictor of offensive labels. Compared to the reference category of ‘very high’ HDI countries, we observed a numerically monotonic increase in offensive labels with decreasing country development: ‘high’: beta = 6.5e-03, 95% CI [-1.34e-04, 0.01], t(1111) = 1.92, p = 0.055; ‘medium’: beta = 0.03, 95% CI [0.02, 0.04], t(1111) = 8.22, p < .001; ‘low’: beta = 0.05, 95% CI [0.04, 0.06], t(1111) = 13.87, p < .001. The monotonic increase in offensive labels with decreasing HDI was confirmed by a Jonckheere-Terpstra test (JT=321394, p=2e-04). This indicates that images of homes from less developed countries were more frequently associated with labels connoting disrepair and death.

**Predicting dCNN misclassifications with country attributes.** To understand why dCNN performance varied widely across countries, we examined dCNN patterns along 13 country attributes from the CIA World Factbook. We fit a linear mixed model to predict mean classification entropy from each country’s attributes, leaving network as a random effect. These findings indicated that countries with higher levels of affluence, education, and urbanization tend to have stronger dCNN classification performance, pointing to the impact of socioeconomic development on computer vision systems' effectiveness. The model's total explanatory power is moderate (conditional $R^2$ = 0.71 with a marginal $R^2$ of 0.06 due to the fixed effects alone), suggesting that factors beyond country attributes influence classification entropy. As shown in Table 4, we found significant negative effects for GDP per capita (beta = -0.41, 95% CI [-0.55, -0.30], t(1271) = -6.58, p<0.001), literacy rate (beta = -0.59, 95% CI [-0.71, -0.48], t(1271) = -9.89, p < .001), education expenditure per capita (beta=0.17, 95% CI [0.07, 0.26], t(1271)=3.41, p<0.001), life expectancy at birth (beta=0.39 95% CI [0.16, 0.62], t(1271)=3.39, p<0.001, maternal mortality rates (beta=-0.18, 95% CI [-0.32, -0.04], t(1271)=-2.45, p=0.014, median age (beta=0.15, 95% CI [0.03, 0.27], t(1271)=2.50, p=0.013, net migration (beta=-0.29, 95% CI [-0.48, -0.11], t(1271)=-3.07, p=0.002, and percent urbanization (beta = -0.16, 95% CI [-0.24, -0.09], t(1271) = -4.52, p<0.001).

| Measure | CR2, MR2 | Intercept | GDP per capita | Population | Literacy | Urbanization | Tourism | Infant mortality | Education expense | Unemployment | Life expectancy | Maternal mortality | Median Age | Area | Net migration |
|---|---|---|---|---|---|---|---|---|---|---|---|---|---|---|---|
| Entropy | **0.71**<br>**0.06** | **3.76**** | **-0.42***** | -0.09 | **-0.59***** | **-0.16***** | -0.07 | -0.10 | **0.17**** | -0.01 | **0.39***** | **-0.18**** | **0.15*** | 0.13 | **-0.29**** |
| Offense | **0.22**<br>**0.14** | **0.11**** | -0.01* | -0.03 | **-0.07***** | -0.01 | -0.01 | -0.03 | 0.007 | -0.007 | -0.002 | -0.006 | **-0.04**** | 0.02 | -0.005 |

**Table 4: Results of linear mixed model analyses predicting dCNN explicit bias for international Airbnb dataset. Beta values are listed in the cells from column 3 onward. Note: predictors were whitened prior to model fitting. Thus, the sign and relative magnitudes are meaningful, but the coefficient values do not straightforwardly translate to specific units. (***: p<0.001; **: p<0.01; *: p<0.05).**

We assessed the extent to which the country attributes could predict the rate of offensive classification labels. These results show that offensive dCNN labels were associated with younger and less literate nations, suggesting that images from older and educated locations dominate the training data for these systems. The model's total explanatory power was moderate (conditional $R^2 = 0.22$, with a marginal $R^2$ of 0.14 due to the fixed effects alone). We found significant negative effects for literacy rate (beta=-0.07 95% CI [-0.09, -0.05], t(1271)=-5.87, p<0.001) and median age (beta=-0.04, 95% CI [-0.06, -0.02], t(1271)=-3.56, p<0.001), see Table 4 for all effects.

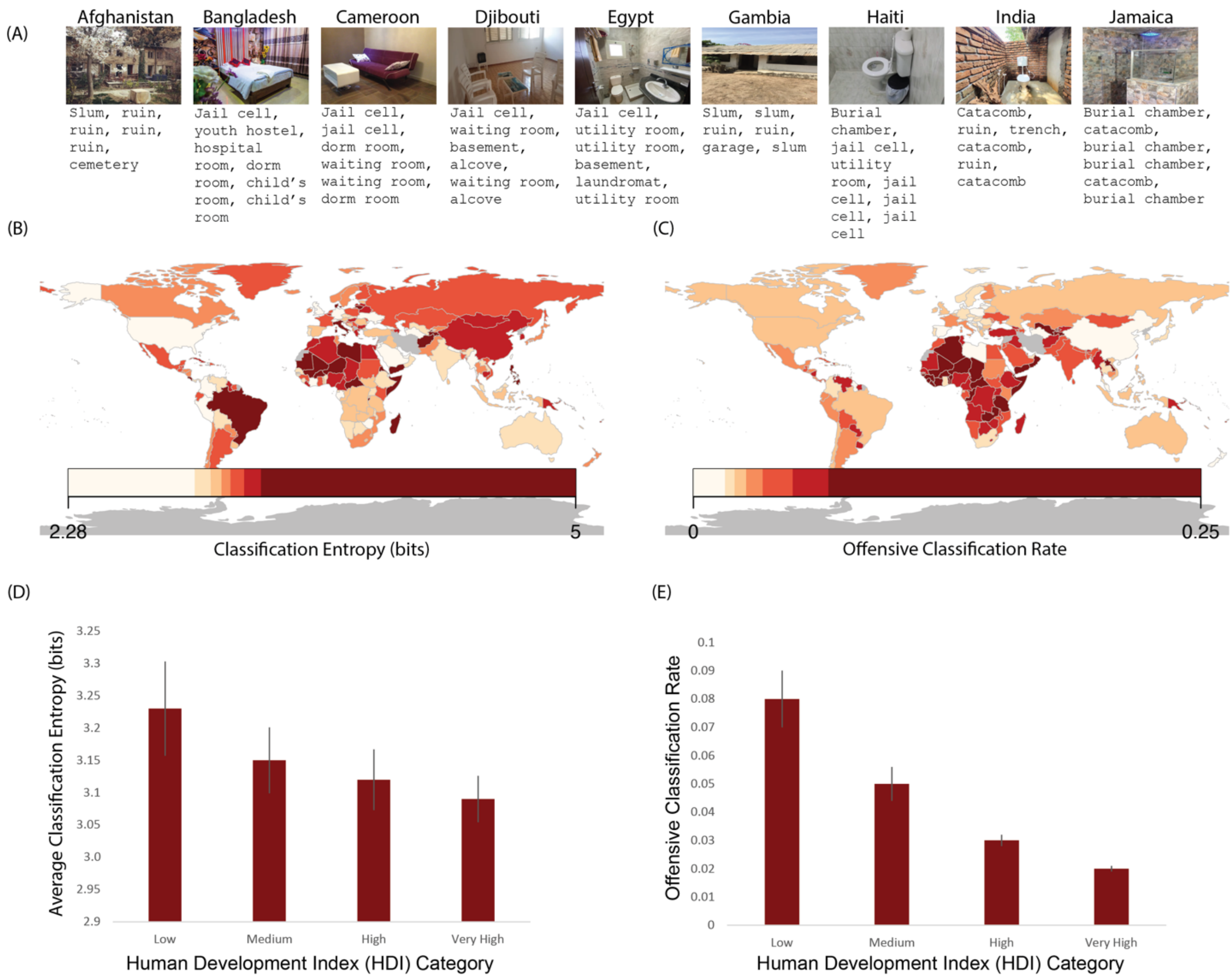

**Figure 2: (A) Examples of bathroom images from Airbnb with sample classifications from the six dCNNs (order: Alexnet, Resnet-18, Resnet-50, Resnet-101, Resnet-152, ConvNext). (B) Map of classification entropy, averaged across the six networks; (C) Map of offensive classification rate, averaged across the six networks; (D) Classification entropy as a function of HDI category level. Error bars reflect 95% confidence intervals; (E) Offensive classification rate as a function of HDI category. Error bars reflect 95% confidence intervals.**

**A multimodal large language model (MLLM) may show less socioeconomic bias than dCNNs.** We assessed the classification accuracy of GPT-4V on this sample. Overall, the accuracy was high (81%). Although there was a numerical tendency for GPT-4V to be more accurate with increasing HDI category (low: 79%, medium: 80%, high: 81%, very high: 82%), this trend did not reach statistical significance (Z=1.06, p=0.14) in the subset of 1000 labeled images. Further, unlike the Dollar Street dataset, we observed no potentially offensive classifications. However, we noticed a marked tendency for GPT-4V to use very vague terms when classifying (e.g., “interior”, “residence”, “room”, etc.) and that this tendency was numerically higher for low HDI countries (11% of terms for low-HDI countries, compared to 7% overall). To further understand this trend, a total of over 40,000 of the images from the Airbnb dataset were classified with GPT-4V, and the number of vague classifications was annotated. A logistic regression revealed that HDI was a significant predictor of a vague classification in the GPT-4V model (beta = -1.35, 95% CI [-1.64, -1.06], $p << .0001$). Therefore, GPT-4V expresses less classification confidence for images with lower socioeconomic status, often hedging towards vague terms for less affluent places.

## Implicit biases in international Airbnb images

The previous results indicated that while dCNNs showed marked socioeconomic biases, the picture for GPT-4V was more subtle. Therefore, we extended the implicit bias analysis employed in the DollarStreet dataset to assess the extent to which implicit bias was present for CLIP for this image set. Using the same methodology, we examined the alignment between each image with both positive and negative concepts. As before, we created a difference score that indicates the net alignment with positive concepts.

**Higher SES photographs are more strongly aligned with positive concepts.** We used ordinary least squares regression to predict the difference score as a function of HDI value for each country. The model explained a statistically significant and moderate proportion of variance ($R^2$ = 0.18, $F(1, 184)$ = 39.96, $p < .001$), with HDI being significantly positively predictive of the difference score (beta = 6.77e-03, 95% CI [4.65e-03, 8.88e-03], $t(184)$ = 6.32, $p << .0001$). This result indicates that photographs from higher HDI countries tend to be more closely associated with positive concepts compared to those from lower HDI countries. This relationship suggests that socioeconomic factors captured by the HDI are reflected in the implicit biases of multimodal deep learning models.

We further investigated the factors contributing to implicit bias by modeling the implicit bias difference score as a function of the socioeconomic and demographic variables from the CIA World Factbook. Our linear model explained a statistically significant and substantial proportion of variance ($R^2$ = 0.30, $F(13, 172)$ = 5.67, $p < .001$, adj. $R^2$ = 0.25). The full results of this analysis are shown in Table 5. Although per-capita GDP was not statistically significant when considering the other factors, we observed that countries with higher positive associations tended to have higher populations (beta = 4.54e-04, 95% CI [1.38e-04, 7.71e-04], $t(172)$ = 2.83, $p$ = 0.005), literacy rates (beta = 4.23e-04, 95% CI [1.07e-04, 7.40e-04], $t(172)$ = 2.64, $p$ = 0.009), more GDP due to tourism (beta = 4.49e-04, 95% CI [1.32e-04, 7.65e-04], $t(172)$ = 2.80, $p$ = 0.006), and higher median age (beta = 7.62e-04, 95% CI [4.45e-04, 1.08e-03], $t(172)$ = 4.75, $p < .001$). At the same time, lower positive associations tended to be found in countries with higher infant mortality rates (beta = -4.19e-04, 95% CI [-7.36e-04, -1.02e-04], $t(172)$ = -2.61, $p$ = 0.010), and higher maternal mortality rates (beta = -5.62e-04, 95% CI [-8.78e-04, -2.45e-04], $t(172)$ = -3.50, $p < .001$). Taken together, these results suggest that lower population health leads to more implicit bias in photographs, while other factors such as literacy and tourism are associated with more positive implicit biases.

| Measure | Intercept | GDP per capita | Population | Literacy | Urbanization | Tourism | Infant mortality | Education expense | Unemployment | Life expectancy | Maternal mortality | Median Age | Area | Net migration |
|---|---|---|---|---|---|---|---|---|---|---|---|---|---|---|
| Implicit bias | **.007*** | -.00007 | **.0005**** | **.0004**** | .0001 | **.0005**** | **-.0004**** | .0002 | -.0002 | .0002 | **-.0006**** | **.0008**** | -.00025 | .0001 |

**Table 5: Results of linear mixed model analyses predicting dCNN implicit bias for international Airbnb dataset. Note: predictors were whitened prior to model fitting. Thus, the sign and relative magnitudes are meaningful, but the coefficient values do not straightforwardly translate to specific units. (***: p<0.001; **: p<0.01; *: p<0.05).**

## Explicit biases in US Airbnb images

Building off our previous analyses that considered international datasets and demonstrated that a country's demographics significantly impact the performance of computer vision models, here we consider the extent to which demographics within a country also modulate machine classification efficacy. This analysis used nearly 800,000 images from all counties in each US state and territory, downloaded from the Airbnb website. It is worth noting that these analyses examine variability within a single country that is classified as "very high" on the Human Development Index scale. As such, we would expect smaller effect sizes compared to the international image sets. However, any significant socioeconomic differences in deep learning systems' performance can be taken as the most rigorous test of socioeconomic bias.

**dCNNs exhibit lower classification accuracy, higher classification uncertainty, and more offensive labels for low SES places within the United States.** We assessed the top-1 and top-5 classification accuracies within our annotated sample of ~1000 images. We fit a logistic mixed model to predict classification accuracy with income percentile as a fixed effect and network architecture as a random effect. We found that increasing income was positively related to increasing top-1 (beta = 1.42, 95% CI [1.19, 1.64], $p < .001$) and top-5 (beta = 1.64, 95% CI [1.37, 1.90], $p < .001$) classification accuracies. Therefore, dCNNs are most accurate with more affluent spaces, even within a single country such as the United States.

We used a linear mixed model to assess the extent to which socioeconomic demographics from the US Census predict the classification performance of deep convolutional neural networks. We found substantial explanatory power in our model (conditional $R^2 = 0.74$; marginal $R^2 = 0.008$). Key predictors such as income, race, and age variability significantly predicted classification entropy (see Figure 3 and Table 6 for all results). Notably, dCNNs showed more certainty in classifying images from counties with higher income, fewer residents of color, and less age diversity. This pattern raises concerns about potential biases in dCNN classifications against certain demographic groups.

We employed a similar model structure in a related analysis examining the rates of offensive label assignment by dCNNs. Despite the overall low rate of offensive classifications (1.4%), our findings revealed meaningful associations: counties with lower health insurance coverage and lower income were more likely to receive offensive labels.

| | Intercept | Health insurance | Income | Educational attainment | Racial diversity | Age diversity |
|---|---|---|---|---|---|---|
| Entropy | **3.00*** ** | -0.004 | **-0.02*** ** | -0.003 | **0.02*** ** | **-0.03*** ** |
| Offensive labels | **0.01*** ** | **-0.0004*** | **-0.007*** ** | -0.0003 | -0.0007 | 0.0003 |

**Table 6: Mixed model coefficients for United States Airbnb set. ***: p<0.001; **: p<0.01; *: p<0.05.**

**A MLLM is more accurate in classifying high SES places within the United States.** Finally, we examined the classification accuracy of GPT-4V for our annotated sample. We found considerable variance in classification accuracy, ranging from 80% for the lower-income counties to 91% for the higher-income counties. As with the dCNNs, we found that increasing income was positively associated with classification accuracy (beta = 1.30, 95% CI [0.62, 2.00], p < .001), indicating that socioeconomic status influenced classification accuracy, even for state-of-the-art models, and even within a single affluent country.

## Implicit biases in US Airbnb images

The previous analyses demonstrated that both dCNNs and GPT-4V exhibited marked explicit socioeconomic biases when classifying Airbnb images from within the United States. As we did with the international datasets, we also wanted to test the extent to which the CLIP model contains implicit biases. We used the same positive and negative concepts as the previous analysis, and created a unified difference score as before.

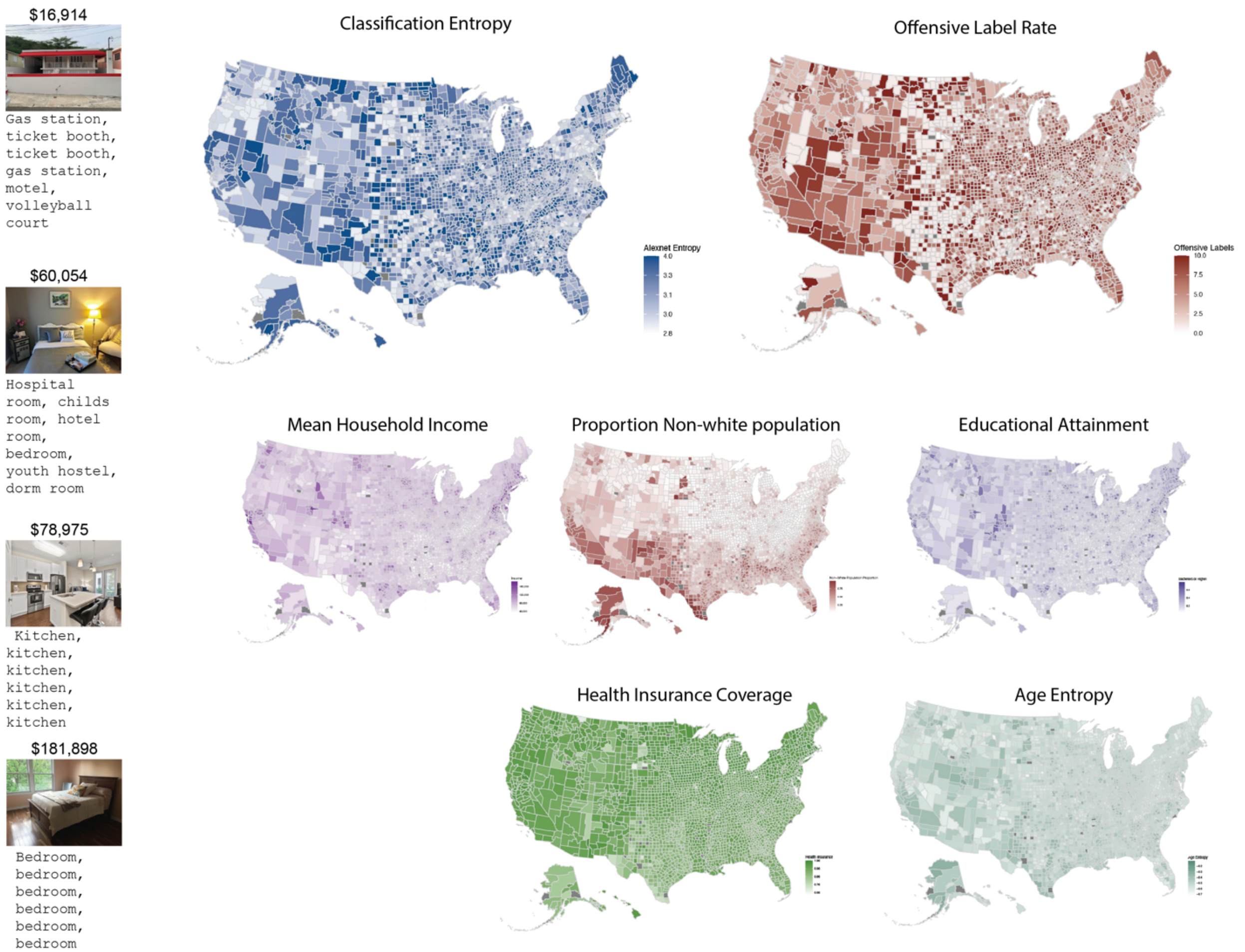

**Figure 3: (Left panel) Example classifications across income levels and networks (order: Alexnet, Resnet-18, Resnet-50, Resnet-101, Resnet-152, ConvNext). (Top row) Classification entropy and offensive label rate (Alexnet shown) projected onto a US county map. (Middle and bottom) Predictor variables from the US Census are projected onto a US county map.**

**Higher SES US locations are better aligned with positive concepts.** We fit a linear model to predict our bias score from the Census variables that were previously described. This statistically significant model explains a moderately weak proportion of variance ($R^2$ = 0.10, F(5, 3137) = 70.13, p < .001, adj. $R^2$ = 0.10). We found that higher household income was associated with stronger positive alignment (beta = 1.65e-04, 95% CI [9.44e-05, 2.36e-04], t(3137) = 4.58, p < .001), as was higher education attainment (beta = 4.03e-04, 95% CI [3.33e-04, 4.74e-04], t(3137) = 11.20, p < .001). These results indicate that images from more educated and affluent counties were better aligned with positive concepts, suggesting a substantial implicit socioeconomic bias within the United States. Interestingly, the proportion of residents of color was positively associated with positive concepts (beta = 3.44e-04, 95% CI [2.73e-04, 4.14e-04], t(3137) = 9.54, p < .001), indicating that there was no substantial racial bias once socioeconomic variables were accounted for. Finally, less age homogeneity was associated with higher positive concept alignment (beta = -3.83e-04, 95% CI [-4.53e-04, -3.12e-04], $t(3137) = -10.62$, $p < .001$). Taken together, areas with greater economic prosperity, educational attainment, racial diversity, and age diversity are more likely to be associated with positivity on a conceptual level by a CLIP model.

# Discussion

Using nearly one million images of homes across three datasets, we found that computer vision models (deep convolutional neural networks and a multimodal large language model) demonstrated lower classification accuracy, lower classification confidence, and a higher propensity to assign potentially offensive labels to images from homes with lower socioeconomic status. While the effect was most dramatic when comparing international image sets, we found that economic and racial variability by US county also predicted classification performance. Moreover, another multimodal model showed substantial implicit biases in all three datasets, consistently showing increased alignment between positive concepts and images from affluent locations. These findings highlight crucial socioeconomic and developmental factors influencing computer vision performance and raise important considerations about the potential for unintended biases in automated image categorization systems.

This research adds to the growing body of literature on biases present in various forms of artificial intelligence. Word embedding models, trained on large text corpora such as Wikipedia, often reinforce stereotypical assumptions about different groups. For instance, these models are more likely to link women with secretarial roles than with computer programming and homosexual men with hairdressing rather than scientific research [25–27]. Furthermore, image search results tend to perpetuate gender stereotypes: searches for adjectives like “ambitious” or “determined” predominantly return images of men, while “emotional” and “insecure” yield pictures of women [3]. Image captioning models tend to describe white men more accurately and diversely than other groups [4]. They also display biases, such as categorizing all images of Black athletes as basketball players regardless of the actual sport [28] and associating women more with wedding dress stores than mechanic shops [5,6]. Similar biases have been observed in visual question-and-answer systems [7]. Troublingly, image classification systems show less efficacy in detecting people with darker skin tones [1] or images of females and children [2] as human beings. Generative image models, such as Stable Diffusion [29], generate images that align with common gender and race stereotypes [8]. For example, the prompt “attractive person” yielded all images of people with light skin tones, and the prompt “flight attendant” yielded only images of women.

By contrast, our study examines whether these biases extend to the homes where people live. One related study examining the Dollar Street dataset found that object classification rates were lower for objects from lower-income families [30]. Our study corroborates and extends these findings. First, it further explains these failures by fitting models specifically designed to predict model classification performance with demographic data. Second, it includes more powerful, transformer-based architectures. Third, it assesses the implicit biases of these models using a tool inspired by social psychology.

The performance of computer vision systems diminishes when classifying images that differ from those they were trained on [31,32]. Extant training sets have been shown to overrepresent images in North America and Europe [30,33]. While these results would have predicted our findings in the international image sets, they do not fully account for the patterns of performance observed within the United States. Instead, our results highlight the need for more inclusive training datasets that better represent global and socioeconomic diversity within countries. Nor do these biases disappear when considering state-of-the-art multimodal large language models (MLLMs) such as GPT-4V, as this model showed significant socioeconomic biases in all three datasets. Moreover, another multimodal model, Contrastive Language-Image Pre-Training (CLIP), showed substantial implicit biases by more strongly aligning positive concepts such as “joy” and “freedom” with more affluent locations and strong alignment between negative concepts like “sickness” and “grief” with lower SES locations.

While dCNNs classify from a fixed set of category labels, MLLMs do not share this constraint. Interestingly, GPT-4V tended to use incredibly vague terms like “interior” or “room” for lower SES locations. There may be multiple reasons for this behavior, including hedging against classification uncertainty or a fairness-related constraint introduced by engineers to reduce potentially offensive classifications. Regardless of origin, the effect sizes reported here underestimate actual bias in MLLMs. The implicit bias analyses corroborate this. That computer vision models, like human observers [24], have implicit biases may be telling. Even within the narrow scope of United States counties, we observed substantial implicit bias that more strongly associated affluent and educated locations with positive concepts.

The amplification of weak correlations within datasets is an unfortunate consequence of algorithmic bias [4,8,34], and these biases are increasing [6]. Further, the seemingly objective nature of AI systems often masks the underlying biases [35,36]. We anticipate several negative consequences from unequal classification performance in computer vision systems. First, as image data are beginning to be used automatically to assess home values [10,11], the race-based bias we observed may lead to a “digital redlining” of homes in non-white neighborhoods. Other efforts have noted that visual scene data can predict crime rates and voting patterns [37,38]. When AI-driven analyses of residential areas are used for automated policing or urban planning, biased scene classification can lead to misinformed policies or initiatives that do not adequately address the needs of diverse communities. If an AI system disproportionately identifies certain neighborhoods as 'run-down' or 'unsafe' based on biased scene classifications, it could unfairly influence home loans, funding allocation, maintenance, and development decisions. Finally, inequities in home recognition may also lead to unequal access to working smart home technologies.

The increased likelihood of assigning offensive labels to images from less developed countries or less diverse US counties may reflect cultural biases embedded within the AI systems, necessitating a reevaluation of the training and development processes to ensure fairness and inclusivity. Several recent works have identified issues at many steps of this development pipeline. Offensive materials are frequently found in large datasets [13–15]. Human observers provide ground truth labels, so these human biases get embedded into AI systems [16]. Further, image datasets are convenience samples and thus overrepresent North America and Europe [33] due to online cultural hegemony [39]. Because stereotypes can emerge from the search results themselves [3,40], creating datasets through web scraping amplifies biases. Finally, these geographic and economic biases may go unnoticed because there is insufficient diversity in the field of artificial intelligence [41] and because those with social biases do not notice them [24,42].

In conclusion, our analysis across three diverse image datasets highlights the pervasive issue of socioeconomic bias in computer vision models used for image classification. These biases manifest explicitly as lower accuracy, confidence, and an increased likelihood of assigning offensive labels to images from lower socioeconomic backgrounds and implicitly as higher alignment between positive concepts and affluent locations. This issue is not confined to international comparisons but is also evident within the United States, underscoring the need for more inclusive and representative training datasets that capture global and within-country socioeconomic diversity. The tendency of AI systems to perpetuate and amplify existing societal biases is a significant concern. The risk of digital redlining, misinformed urban planning, and unfair allocation of resources based on biased AI assessments of residential areas is serious and imminent. These findings call for a critical reevaluation of AI development processes, from dataset creation to algorithm training, to ensure fairness and inclusivity. This reevaluation must address the issues at various stages of AI development, including the inherent biases in large datasets, the cultural biases in ground truth labels, the overrepresentation of certain geographies in training data, and the lack of diversity in the AI field. Addressing these challenges is crucial for developing equitable AI systems that do not inadvertently perpetuate systemic inequalities.

# Methods

## Deep learning systems

For each dataset, we report the efficacy of six dCNN architectures (AlexNet [43], Resnet-18 [44], Resnet-50 [44], Resnet-101 [44], Resnet-152 [44], and ConvNext [45]. The first three architectures were pretrained on the Places-365 dataset [12]. These images were scraped from the web using several popular search engines, including Google Images, Bing Images, and Flikr. The final three architectures were pretrained on the ImageNet dataset and fine-tuned on Places-365. We froze the weights in all layers except the last fully-connected layer and changed the output layer size from 1000 to 365, reflecting the number of scene categories in the Places dataset. This approach allowed us to retain the basic visual features learned from ImageNet while allowing the fully-connected layers to specialize for features necessary for scene classification.

We employed GPT4-Vision preview (OpenAI), a multimodal large language model (MLLM) capable of providing text outputs to image inputs to capture state-of-the-art machine learning capabilities. Images were encoded into base64 and sent to the OpenAI API with the following prompt: “What is the basic-level category of this place? Use a single category label.” The resulting predictions were collated and assessed for accuracy, potential offensiveness, and overly vague labels. To test implicit biases, we used Contrastive Language-Image Pre-Training (CLIP, ViT-B/32, OpenAI) [17].

## Stimuli

### Dollar Street

We used the Dollar Street dataset to obtain image examples from real living situations [18]. This dataset was created by photographing 264 homes in 54 countries and consists of 135 classes that include objects (e.g., "alarm clock"), places (e.g., "bathroom"), events (e.g., "adding spices to food when cooking"), people (e.g., "families"), and abstract categories (e.g., "favorite home decorations"). From these classes, we selected the four that described scene categories in the Places database [12]: bathroom, bedroom, kitchen, and living room. Home exteriors were also downloaded, but as they could map onto multiple Places categories, they were omitted from this analysis. We downloaded a random sample of up to ten images per country to ensure that heavily-represented countries did not skew continent-level results. We included a total of 1193 images in the analysis. The Dollar Street dataset also contains the monthly monetary consumption of each household, reported in US dollars and adjusted for purchasing power parity (PPP). A detailed description of how this metric was computed can be found on the website [18]. Images were collected in the Summer of 2023.

### Airbnb

To get a wider representation of homes and countries, we scraped the website Airbnb for images of family homes in 219 countries and territories in the spring of 2021. A list of countries and territories can be found in the Supplementary Information. Our scraper downloaded the first five images for each property that was listed in each country. Properties were excluded if they were hotels or in the top half of the cost distribution for a given region. This choice was designed to ensure that images reflected the lived experience of family life, rather than investment properties. Images were inspected by hand and omitted if they did not depict a home scene category (for example, family portraits, area maps, or local scenic attractions), or if they contained a person. The number of images retained per country ranged from one in the Pitcairn Islands and Yemen to 1334 in Montenegro. A total of 118,690 images were included in the analysis.

To estimate classification accuracy, 12 countries were selected (three countries randomly selected from each of the four HDI categories: (Antigua and Barbuda (high), Central African Republic (low), Iceland (very high), Jordan (high), Norway (very high), Romania (high), Somalia (low), South Sudan (low), Switzerland (very high), Tonga (medium), Tunisia (medium), Uruguay (high)). For each country, up to 100 images were randomly selected for annotation, for a total of 887 images. The first author matched each image to the closest scene category from the Places dataset.

We applied similar methods to the 3220 United States counties in US states and territories. To ensure that properties lay within each county line, we created a custom URL for each county with map borders adjusted by hand. As with the international dataset, we downloaded five images from each property and applied the same exclusion criteria. A total of 795,856 images were included in the analysis. We similarly selected seven United States counties to annotate ground truth categories. The counties were selected to span the range of mean income from the 2020 US Census: Guánica Municipio, Puerto Rico; Hudspeth County, Texas; Caldwell County, North Carolina; Cumberland County, Virginia; Duval County, Florida; Charles County, Maryland; Falls Church County, Virginia. Up to 100 images were randomly selected per county and annotated by the first author (total of 570 images).

## Covariates

To predict the relative success of dCNN classifications for the international datasets, we aggregated the following information from the CIA World Factbook [46]: the area of each country and region (in $km^2$), GDP-per-capita (in USD), population (in millions), literacy rate, the proportion of urbanized land, infant mortality rate, education expenditure percentage of GDP, youth unemployment rate, life expectancy at birth, maternal mortality rate, and net migration rate. We omitted eight regions that were missing more than three of these variables (Christmas Island, Falkland Islands, Holy See, Jay Mayen, Niue, Norfolk Islands, Pitcairn Islands, Saint Barthelemy). When possible, we filled in missing data from the World Bank and UNESCO for the remaining countries. For any remaining missing values, we used a K-nearest neighbor approach to data imputation whereby we computed the Euclidean distance between the country's feature vector and the feature vectors of all countries with full descriptions (omitting the missing variable from the comparison countries). We estimated missing values as the average variable value for the three most similar countries. All variables were normalized between 0 and 1 prior to analysis. As some of these predictors were highly correlated, we applied ZCA whitening to the predictor matrix, as suggested by [47] to reduce multicollinearity in our regression models.

To predict the relative classification success for the United States dataset, we extracted the following variables from each county from the 2020 United States Census American Community Survey [48]: Age and Sex (S0101), from which we computed the age entropy of each county; Educational Attainment (S1501), from which we calculated the proportion of each county's residents with a bachelor's degree or higher; Mean Income (S1902); Health Insurance Coverage (S2701); and Race and Ethnicity (S0201), from which we computed the proportion of each county's residents who are not white.

## Dependent measures

When possible, we computed three measures of classification success for the dCNNs. A classification was considered top-1 accurate if the most probable category in the network's final layer corresponded to the correct category. A classification was top-5 accurate if any of the five most probable categories corresponded to the true classification. These metrics are widely used in computer vision to account for noise in ground truth labels [49]. We employed these metrics for the Dollar Street images, as these had ground truth labels. For the Airbnb datasets, we employed the metric of classification entropy. As the activations of the last layer of a deep network can be considered a discrete probability distribution over scene categories, computing entropy on this distribution [50] shows us the relative uncertainty of the network's classification. For example, as the Places dataset contains 365 unique scene categories, the maximum possible entropy is $\log_2(365)$ or ~8.5 bits, corresponding to an assessment of each scene category being equally likely (i.e., maximal confusion). Classification entropy will approach zero bits for a network that classifies one category with probability ~1.

## Implicit bias measurement

Images were preprocessed using CLIP's standard pipeline. We collated a set of 20 positive concepts ('joy', 'love', 'peace', 'wonderful', 'pleasure', 'friend', 'laughter', 'happy', 'freedom', 'health', 'cheer', 'gentle', 'honest', 'gift', 'miracle', 'paradise', 'safe', 'vacation', 'beautiful', 'healthy') and 20 negative concepts ('abuse', 'crash', 'filth', 'murder', 'sickness', 'accident', 'death', 'grief', 'poison', 'stink', 'assault', 'disaster', 'hatred', 'pollute', 'tragedy', 'bomb', 'poverty', 'rotten', 'ugly', 'jail'). Each image in the dataset was encoded using CLIP to extract its visual features. Then, each of the 40 concept words was tokenized and encoded using the same model to generate textual embeddings. We evaluated image-concept alignment by computing the normalized cosine similarity between image and word embeddings. For each image, we averaged the similarity scores for all positive concepts and the similarity scores for all negative concepts. We computed the positive-negative difference, and this difference score was our measure of implicit bias.

## Ethical Approval

This article does not contain any studies with human participants performed by any of the authors.

## Informed Consent

This article does not contain any studies with human participants performed by any of the authors.

## Conflicts of Interest

The authors declare no competing interests.

## Author Contributions

MRG conceived of the research, which was conducted by MRG, MJ, WS, and JAH. MRG wrote the manuscript with editorial input from MJ, WS, and JAH.

## Data Availability

All data, stimuli, and scripts are available here: https://osf.io/4vqu8/.

# Acknowledgments

The authors thank Alexis Fifield, Noah Schwartz, and Adam Katz for helpful conversations about this manuscript.